\documentclass[10pt,twocolumn,letterpaper]{article}

\usepackage{iccv}
\usepackage{times}
\usepackage{epsfig}
\usepackage{graphicx}
\usepackage{amsmath}
\usepackage{amssymb}

\usepackage{multirow}
\usepackage{textcomp}
\usepackage{subfiles}
\usepackage{adjustbox}
\usepackage{siunitx}
\usepackage[inline, shortlabels]{enumitem}

\DeclareMathOperator*{\argmin}{arg\,min}
\newcommand{\ma}[1]{\mathrm{#1}}

\newcommand{\norm}[1]{\left\lVert#1\right\rVert}
\newcommand{\st}{\text{subject to}}

\newcommand{\round}[2]{\num[round-mode=places,round-precision=#1]{#2}}
\newcommand{\stitle}[1]{\noindent\textbf{#1}\\}
\newcommand{\lcoef}{\boldsymbol{\beta}}
\newcommand{\lcoefi}{\beta_i}
\newcommand{\lcoefx}[1]{\beta_#1}
\newcommand{\wtwo}{\ma{W}}
\newcommand{\wtwoi}{\ma{W_i}}
\newcommand{\wtwox}[1]{\ma{W_#1}}
\newcommand{\reW}{\ma{W^\prime}}
\newcommand{\reY}{\ma{Y}}
\newcommand{\reX}{\ma{X}}
\newcommand{\reXi}{\ma{X_i}}
\newcommand{\reXx}[1]{\ma{X_#1}}
\newcommand{\lalpha}{\lambda}
\newcommand{\rank}{c}
\newcommand{\out}{n}
\newcommand{\kh}{k_h}
\newcommand{\kw}{k_w}
\newcommand{\samp}{N}
\newcommand{\rankp}{c^\prime}

\newcommand{\wxi}{\ma{Z_i}}
\newcommand{\cx}{\ma{X^\prime}}

\newcommand{\x}[1]{$#1\times$}

\newcommand{\las}{LASSO regression}
\newcommand{\conv}{convolutional layer}
\newcommand{\featch}{feature map width}

\newcommand{\ratio}{speed-up ratio}

\newcommand{\firstk}{\textit{first k}}
\newcommand{\prune}{\textit{max response}}

\newcommand{\filterwise}{\textit{filter-wise pruning}}
\newcommand{\Filterwise}{\textit{Filter-wise pruning}}

\newcommand{\sampling}{\textit{feature map sampling}}

\newcommand{\filterpruning}{Filter pruning}
\newcommand{\tensordecom}{tensor factorization}
\newcommand{\Tensordecom}{Tensor factorization}

\newcommand{\channelpruning}{channel pruning}
\newcommand{\Channelpruning}{Channel pruning}

\newcommand{\structured}{structured simplification}
\newcommand{\Structured}{Structured simplification}
\newcommand{\implementation}{optimized implementation}
\newcommand{\Implementation}{Optimized implementation}
\newcommand{\sparseconnect}{sparse connection}
\newcommand{\Sparseconnect}{Sparse connection}

\newcommand{\origvgg}{89.9}
\newcommand{\vggtworaw}{2.7}
\newcommand{\vggfour}{1.0}
\newcommand{\vggfouracc}{11.1} 
\newcommand{\vggfourraw}{7.9}
\newcommand{\vggfourrawacc}{18.0} 
\newcommand{\vggfive}{1.7}
\newcommand{\vggfiveraw}{22.0}
\newcommand{\vggscratchacc}{11.9} 
\newcommand{\vggscratcherr}{1.8}
\newcommand{\vggscratchuniacc}{12.5}
\newcommand{\vggscratchunierr}{2.4}

\newcommand{\vggc}{1.3}
\newcommand{\vggcft}{0.3}
\newcommand{\vggcfour}{0.7} 
\newcommand{\vggcftfour}{0.0}

\newcommand{\resorig}{92.2} 
\newcommand{\resmb}{4.0} 

\newcommand{\resft}{1.4}

\newcommand{\xceptionfifty}{Xception-50}
\newcommand{\xceptionpr}{4.3}
\newcommand{\xceptionorig}{92.8}
\newcommand{\xceptioncr}{2.9}
\newcommand{\xceptionft}{1.0}

\usepackage[pagebackref=true,breaklinks=true,letterpaper=true,colorlinks,bookmarks=false]{hyperref}

\iccvfinalcopy 

\def\iccvPaperID{600} 

\begin{document}

\title{Channel Pruning for Accelerating Very Deep Neural Networks}

\author{Yihui He\thanks{This work was done when Yihui He was an intern at Megvii Inc.}\\
	Xi'an Jiaotong University\\
	Xi'an, 710049, China\\
	{\tt\small heyihui@stu.xjtu.edu.cn}
	\and
	Xiangyu Zhang\\
	Megvii Inc.\\
	Beijing, 100190, China\\
	{\tt\small zhangxiangyu@megvii.com}
	\and
	Jian Sun\\
	Megvii Inc.\\
	Beijing, 100190, China\\
	{\tt\small sunjian@megvii.com}
}

\maketitle

\begin{abstract}
	In this paper, we introduce a new channel pruning method to accelerate very deep convolutional neural networks. 
	Given a trained CNN model, we propose an iterative two-step algorithm to effectively prune each layer, by a \las\ based channel selection and least square reconstruction. We further generalize this algorithm to multi-layer and multi-branch cases. Our method reduces the accumulated error and enhance the compatibility with various architectures. Our pruned VGG-16 achieves the state-of-the-art results by \x{5} speed-up along with only 0.3\% increase of error. More importantly, our method is able to accelerate modern networks like ResNet, Xception and suffers only \resft\%, \xceptionft\% accuracy loss under \x{2} speed-up respectively, which is significant. Code has been made publicly available\footnote{https://github.com/yihui-he/channel-pruning}.
\end{abstract}

\section{Introduction}

	Recent CNN acceleration works fall into three categories: \implementation\ (\eg, FFT \cite{vasilache2014fast}), 
	quantization (\eg, BinaryNet \cite{courbariaux2016binarynet}), 
	and \structured\ that convert a CNN into compact one \cite{jaderberg2014speeding}.
	This work focuses on the last one.
	
	\begin{figure}
	\centering
	\includegraphics[width=0.9\linewidth]{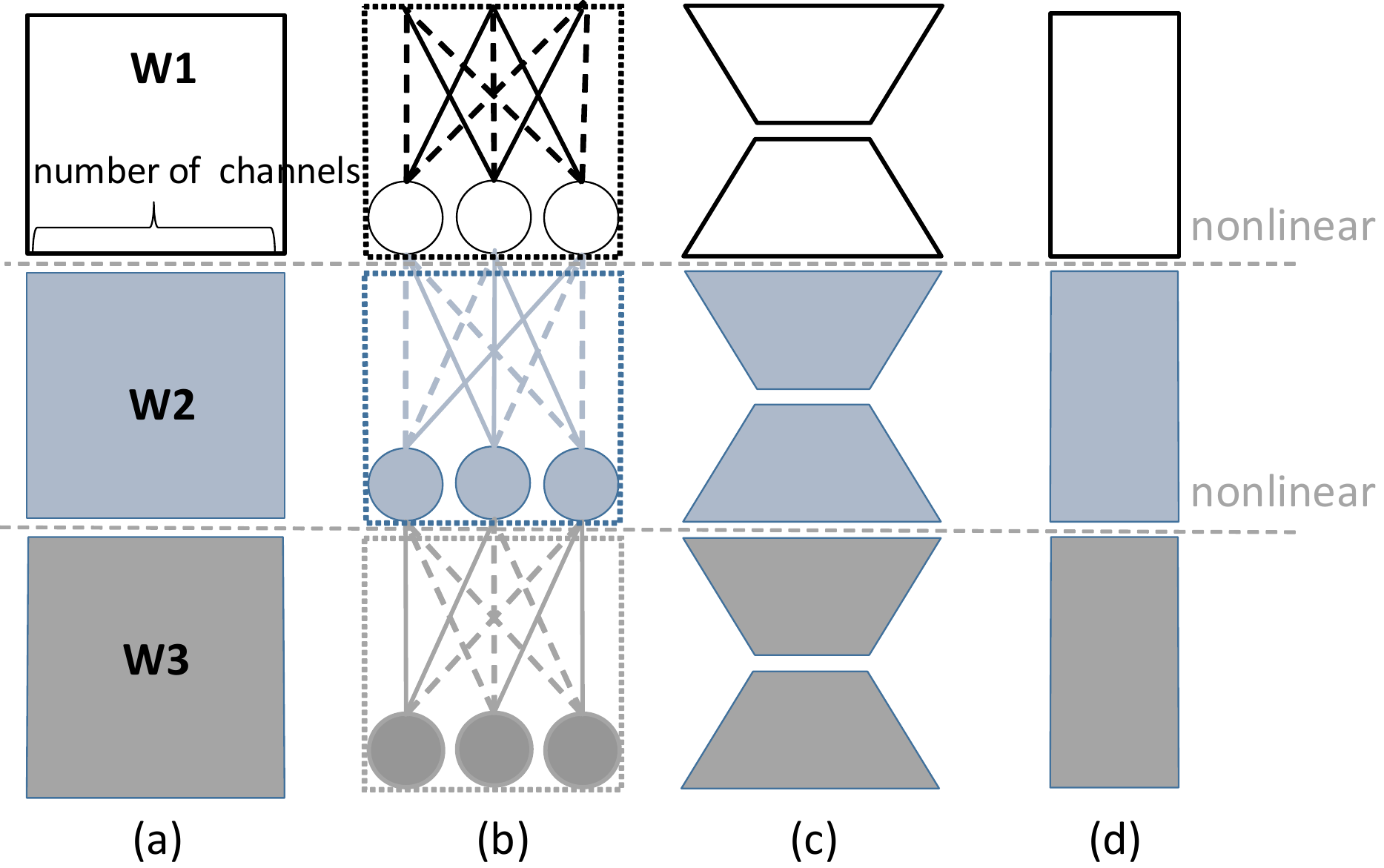}
	\caption{\Structured\ methods that accelerate CNNs: (a) a network with 3 conv layers. (b) \sparseconnect\ deactivates some connections between channels. (c) \tensordecom\ factorizes a \conv\ into several pieces. (d) \channelpruning\ reduces number of channels in each layer (focus of this paper).}
	\label{fig:struct}
	\end{figure}
	\Structured\ mainly involves:
	\tensordecom\ \cite{jaderberg2014speeding},
	\sparseconnect\ \cite{han2015learning},
	and \channelpruning\ \cite{wen2016learning}.
	\Tensordecom\ factorizes a \conv\ into several efficient ones (Fig.~\ref{fig:struct}(c)).
	However, \featch\ (number of channels) could not be reduced,
	which makes it difficult to decompose $1\times1$ \conv\ favored by modern networks (\eg, GoogleNet \cite{szegedy2015going}, ResNet \cite{He2015},  Xception \cite{chollet2016xception}). This type of method also introduces extra computation overhead.
	\Sparseconnect\ deactivates connections between neurons or channels (Fig.~\ref{fig:struct}(b)).
	Though it is able to achieves high theoretical \ratio,
	the sparse \conv s have an "irregular" shape which is not implementation friendly.
	In contrast, \channelpruning\ directly reduces \featch, 
	which shrinks a network into thinner one, 
	as shown in Fig.~\ref{fig:struct}(d).
	It is efficient on both CPU and GPU because no special implementation is required.

	Pruning channels is simple but challenging because removing channels in one layer might dramatically change the input of the following layer. Recently, \textit{training-based} \channelpruning\ works \cite{Alvarez2016,wen2016learning} have focused on imposing sparse constrain on weights during training, which could adaptively determine hyper-parameters.
	However, training from scratch is very costly and results for very deep CNNs on ImageNet have been rarely reported.
	\textit{Inference-time} attempts \cite{Li2016,anwar2016compact} have focused on analysis of the importance of individual weight. The reported \ratio\ is very limited. 

	\begin{figure}
		\centering
		\includegraphics[width=0.8\linewidth]{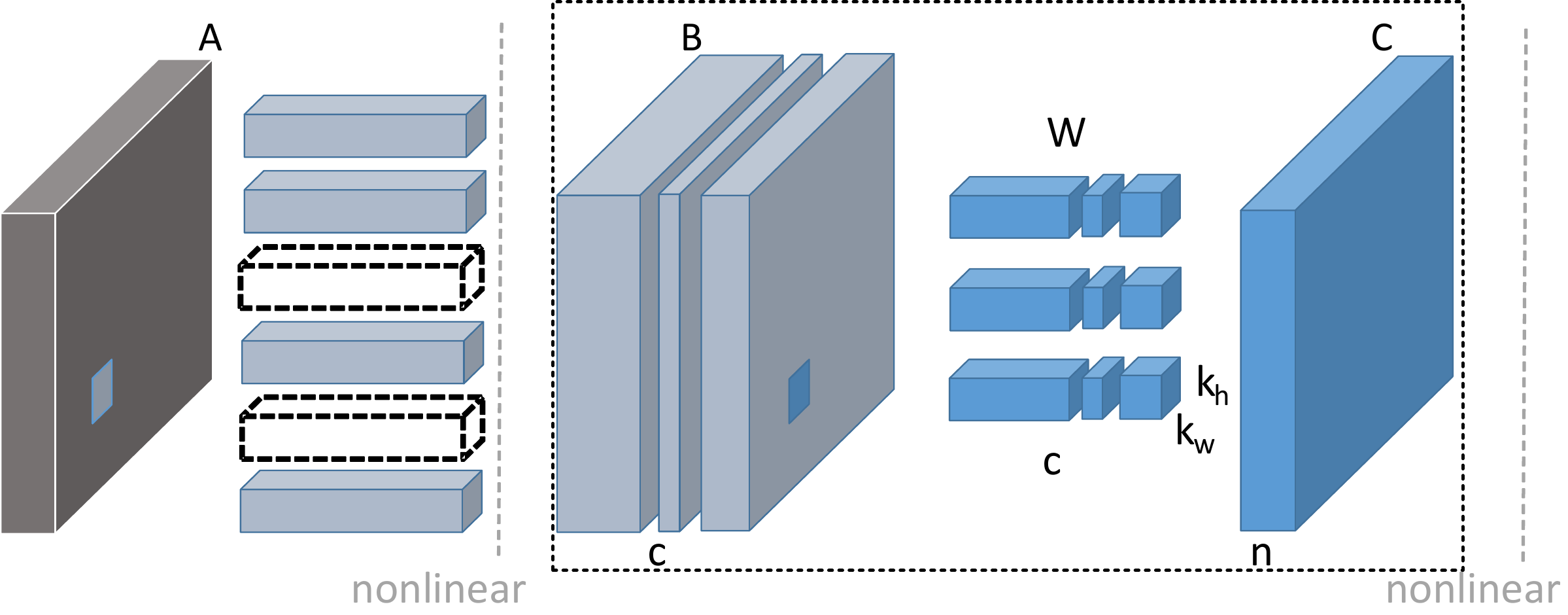}
		\caption{\Channelpruning\ for accelerating a \conv. 
		We aim to reduce the number of channels of feature map B,
		while minimizing the reconstruction error on feature map C.
		Our optimization algorithm (Sec.~\ref{sec:linear}) performs within the dotted box, which does not involve nonlinearity.
		This figure illustrates the situation that two channels are pruned for feature map B. 
		Thus corresponding channels of filters $\wtwo$ can be removed.
		Furthermore, even though not directly optimized by our algorithm, 
		the corresponding filters in the previous layer can also be removed
		(marked by dotted filters).
		 $c,n$: number of channels for feature maps B and C, $k_h\times k_w$: kernel size. 
		}
		\label{fig:ill}
	\end{figure}

	In this paper, we propose a new inference-time approach for \channelpruning, 
	utilizing redundancy inter channels.
	Inspired by \tensordecom\ improvement by feature maps reconstruction \cite{Zhang2015}, 
	instead of analyzing filter weights \cite{jaderberg2014speeding,Li2016}, 
	we fully exploits redundancy inter feature maps. Specifically, 
	given a trained CNN model, pruning each layer is achieved by minimizing reconstruction error on its output feature maps, 
	as showned in Fig.~\ref{fig:ill}.
	We solve this minimization problem by two alternative steps: channels selection and feature map reconstruction.
	In one step, we figure out the most representative channels,
	and prune redundant ones, based on \las.
	In the other step, we reconstruct the outputs with remaining channels with linear least squares.
	We alternatively take two steps. Further, we approximate the network layer-by-layer, with accumulated error accounted.
	We also discuss methodologies to prune multi-branch networks (\eg, ResNet \cite{He2015}, Xception \cite{chollet2016xception}).
		
	For VGG-16, we achieve \x{4} acceleration, with only \textbf{\vggfour\%} increase of top-5 error.
	Combined with \tensordecom, we reach \x{5} acceleration but merely suffer \textbf{\vggcft\%} increase of error, 
	which outperforms previous state-of-the-arts.
	We further speed up ResNet-50 and Xception-50 by \x{2} with only \textbf{\resft\%, \xceptionft\%} accuracy loss respectively. 

\section{Related Work}

	There has been a significant amount of work on accelerating CNNs.
	Many of them fall into three categories: \implementation\ \cite{bagherinezhad2016lcnn}, 
	quantization \cite{Rastegari2016}, and \structured\ \cite{jaderberg2014speeding}.

	\Implementation\ based methods \cite{mathieu2013fast,vasilache2014fast,lavin2015fast,bagherinezhad2016lcnn} accelerate convolution, with special convolution algorithms like FFT \cite{vasilache2014fast}.
	Quantization \cite{courbariaux2016binarynet,Rastegari2016} reduces floating point computational complexity. 

	\Sparseconnect\ eliminates connections between neurons \cite{han2015learning,liu2015sparse,lebedev2015fast,han2016eie,guo2016dynamic}.
	 \cite{yang2016designing} prunes connections based on weights magnitude. 
	 \cite{han2015deep} could accelerate fully connected layers up to \x{50}.
	However, in practice, the actual speed-up maybe very related to implementation.

	\Tensordecom\ \cite{jaderberg2014speeding,lebedev2014speeding,gong2014compressing,kim2015compression} decompose weights into several pieces.
	\cite{xue2013restructuring,denton2014exploiting,girshick2015fast} accelerate fully connected layers with truncated SVD.
	\cite{Zhang2015} factorize a layer into $3\times3$ and $1\times1$ combination, driven by feature map redundancy.

	\Channelpruning\ removes redundant channels on feature maps.
	There are several training-based approaches. 
	\cite{Alvarez2016,wen2016learning,zhou2016less} regularize networks to improve accuracy.
	Channel-wise SSL \cite{wen2016learning} reaches high compression ratio for first few conv layers of LeNet \cite{lecun1998gradient} and AlexNet \cite{krizhevsky2012imagenet}.
	\cite{zhou2016less} could work well for fully connected layers.
	However,
	\textit{training-based} approaches are more costly,
	and the effectiveness for very deep networks on large datasets is rarely exploited.
	
	
	Inference-time \channelpruning\ is challenging, as reported by previous works \cite{anwar2015structured,polyak2015channel}.
	Some works \cite{srinivas2015data,mariet2015diversity,hu2016network} focus on model size compression, which mainly operate the fully connected layers.
	Data-free approaches \cite{Li2016,anwar2016compact} results for \ratio\ (\eg, \x{5}) have not been reported,
	and requires long retraining procedure.
	 \cite{anwar2016compact} select channels via over 100 random trials, 
	however it need long time to evaluate each trial on a deep network,
	which makes it infeasible to work on very deep models and large datasets.
	\cite{Li2016} is even worse than naive solution from our observation sometimes (Sec. \ref{sec:ablation}).
	

\section{Approach}
	
	In this section, we first propose a \channelpruning\ algorithm for a single layer,
	then generalize this approach to multiple layers or the whole model.
	Furthermore, we discuss variants of our approach for multi-branch networks.
	
	\subsection{Formulation}\label{sec:linear}
	\newcommand{\rankcontrain}{ \norm{\lcoef}_0 \leq \rankp}
	\newcommand{\wtwocontrain}{ \forall i \norm{\wtwoi}_F =1}	
	
	Fig.~\ref{fig:ill} illustrates our \channelpruning\ algorithm for a single \conv.
	We aim to reduce the number of channels of feature map B, while maintaining outputs in feature map C.
	Once channels are pruned, 
	we can remove corresponding channels of the filters that take these channels as input.
	Also, filters that produce these channels can also be removed.
	It is clear that \channelpruning\ involves two key points. 
	The first is channel selection, since we need to select proper channel combination to maintain as much information.
	The second is reconstruction. We need to reconstruct the following feature maps using the selected channels.
	
	Motivated by this, we propose an iterative two-step algorithm.
	In one step, we aim to select most representative channels.
	Since an exhaustive search is infeasible even for tiny networks,
	we come up with a \las\ based method to figure out representative channels and prune redundant ones.
	In the other step, we reconstruct the outputs with remaining channels with linear least squares.
	We alternatively take two steps.
	
	Formally, to prune a feature map with $\rank$ channels, we consider applying $\out \times \rank \times \kh \times \kw$ convolutional filters $\wtwo$ on $\samp \times \rank  \times \kh \times \kw$ input volumes $\reX$ sampled from this feature map, which produces $\samp \times \out$ output matrix $\reY$.
	Here, $\samp$ is the number of samples, $\out$ is the number of output channels, and $\kh, \kw$ are the kernel size. 
	For simple representation, bias term is not included in our formulation.
	To prune the input channels from $\rank$ to desired $\rankp$ ($0 \leq \rankp \leq \rank$), while minimizing reconstruction error, 
	we formulate our problem as follow:
	\begin{equation}\label{eq:l0}
	\begin{aligned}
	& 	\argmin_{\lcoef, \wtwo} \frac{1}{2\samp} \norm{\reY - \sum_{i=1}^{\rank} \lcoefi \reXi \wtwoi^\top }^2_F \\
	& \st \rankcontrain
	\end{aligned}
	\end{equation}
	
	$\norm{ \cdot }_F$ is Frobenius norm.
	$\reXi$ is $\samp \times \kh\kw$ matrix sliced from $i$th channel of input volumes $\reX$, $i = 1,...,\rank$.
	$\wtwoi$ is $\out \times \kh\kw$ filter weights sliced from $i$th channel of $\wtwo$. 
	$\lcoef$ is coefficient vector of length $\rank$ for channel selection, and $\lcoefi$ ($i$th entry of $\lcoef$) is a scalar mask to $i$th channel (i.e. to drop the whole channel or not).
	Notice that, if $\lcoefi = 0$, $\reXi$ will be no longer useful,
	which could be safely pruned from feature map.
	$\wtwoi$ could also be removed.
	$\rankp$ is the number of retained channels, which is manually set as it can be calculated from the desired speed-up ratio. For whole-model speed-up (i.e. Section \ref{sec:ratio}), given the overall speed-up, we first assign speed-up ratio for each layer then calculate each $\rankp$.
	
	\stitle{Optimization}
	Solving this $\ell_0$ minimization problem in Eqn.~\ref{eq:l0} is NP-hard. 
	Therefore, we relax the $\ell_0$ to $\ell_1$ regularization:
	\begin{equation}\label{eq:l1}
	\begin{aligned}
	& 	\argmin_{\lcoef, \wtwo} \frac{1}{2\samp} \norm{\reY - \sum_{i=1}^{\rank} \lcoefi \reXi \wtwoi^\top }^2_F + \lalpha \norm{\lcoef}_1 \\
	& \st \rankcontrain, \wtwocontrain
	\end{aligned}
	\end{equation}
	
	$\lalpha$ is a penalty coefficient.
	By increasing $\lalpha$, there will be more zero terms in $\lcoef$ and one can get higher \ratio.
	We also add a constrain $\wtwocontrain$ to this formulation, 
	which avoids trivial solution.

	Now we solve this problem in two folds.
	First, we fix $\wtwo$, solve $\lcoef$ for channel selection.	
	Second, we fix $\lcoef$, solve $\wtwo$ to reconstruct error.
	
	\textbf{(i) The subproblem of $\lcoef$}. 
	In this case, $\wtwo$ is fixed. 
	We solve $\lcoef$ for channel selection.
	This problem can be solved by \las\ \cite{tibshirani1996regression,breiman1995better}, which is widely used for model selection.
	\begin{equation}
	\begin{aligned}
	& \hat{\lcoef}^{LASSO}(\lalpha)= \argmin_{\lcoef} \frac{1}{2\samp}\norm{\reY -  \sum_{i=1}^{\rank} \lcoefi\wxi}_F^2 + \lalpha \norm{\lcoef}_1 \\
	& \st \rankcontrain
	\end{aligned}
	\end{equation}
	Here $\wxi = \reXi\wtwoi^\top$ (size $\samp \times \out$).
	We will ignore $i$th channels if $\lcoefi = 0$.
	
	\textbf{(ii) The subproblem of $\wtwo$}.
	In this case, $\lcoef$ is fixed.
	We utilize the selected channels to minimize reconstruction error.
	We can find optimized solution by least squares:
	\begin{equation}
	\argmin_{\reW} \norm{\reY - \cx(\reW)^\top}_F^2
	\end{equation}
	Here $\cx = [\lcoefx{1}\reXx{1}\; \lcoefx{2}\reXx{2}\; ...\; \lcoefx{i}\reXx{i}\; ...\; \lcoefx{c}\reXx{c}]$ (size $\samp \times  \rank\kh\kw$). 
	$\reW$ is $\out \times \rank\kh\kw$ reshaped $\wtwo$, 
	$\reW=[\wtwox{1}\; \wtwox{2}\; ...\; \wtwox{i}\; ...\; \wtwox{c}]$.
	After obtained result $\reW$, it is reshaped back to $\wtwo$.
	Then we assign $\lcoefi \leftarrow \lcoefi \norm{\wtwoi}_F, \wtwoi \leftarrow \wtwoi / \norm{\wtwoi}_F$. 
	Constrain $\wtwocontrain$ satisfies.

	We alternatively optimize (i) and (ii).
	In the beginning, $\wtwo$ is initialized from the trained model,  $\lalpha=0$, namely no penalty,
	and $\norm{\lcoef}_0=\rank$.
	We gradually increase $\lalpha$. 
	For each change of $\lalpha$, we iterate these two steps until $\norm{\lcoef}_0$ is stable. 
	After $\rankcontrain$ satisfies, 
	we obtain the final solution $\wtwo$ from $\{\lcoefx{i}\wtwox{i}\}$.
	In practice, we found that the two steps iteration is time consuming. 
	So we apply (i) multiple times, until $\rankcontrain$ satisfies.
	Then apply (ii) just once, to obtain the final result.
	From our observation, this result is comparable with two steps iteration's.
	Therefore, in the following experiments, we adopt this approach for efficiency.
	

	\textbf{Discussion}:
	Some recent works \cite{wen2016learning,Alvarez2016,han2015learning} (though training based) also introduce $\ell_1$-norm or LASSO.
	However, we must emphasis that we use different formulations.
	Many of them introduced sparsity regularization into training loss, 
	instead of explicitly solving LASSO.
	Other work \cite{Alvarez2016} solved LASSO, while feature maps or data were not considered during optimization.
	Because of these differences, our approach could be applied at inference time.
	
	\subsection{Whole Model Pruning}\label{sec:whole}
	Inspired by \cite{Zhang2015}, we apply our approach layer by layer sequentially.
	For each layer, we obtain input volumes from the current input feature map, 
	and output volumes from the output feature map of the un-pruned model.
	This could be formalized as: 

	\begin{equation}\label{eq:whole}
	\begin{aligned}
	& 	\argmin_{\lcoef, \wtwo} \frac{1}{2\samp} \norm{\reY^\prime - \sum_{i=1}^{\rank} \lcoefi \reXi \wtwoi^\top }^2_F \\
	& \st \rankcontrain
	\end{aligned}
	\end{equation}
	
	Different from Eqn.~\ref{eq:l0}, $\reY$ is replaced by $\reY^\prime$, 
	which is from feature map of the original model.
	Therefore, the accumulated error could be accounted during sequential pruning.
	
	\subsection{Pruning Multi-Branch Networks}\label{sec:variants}
\begin{figure}
	\centering
	\includegraphics[width=0.8\linewidth]{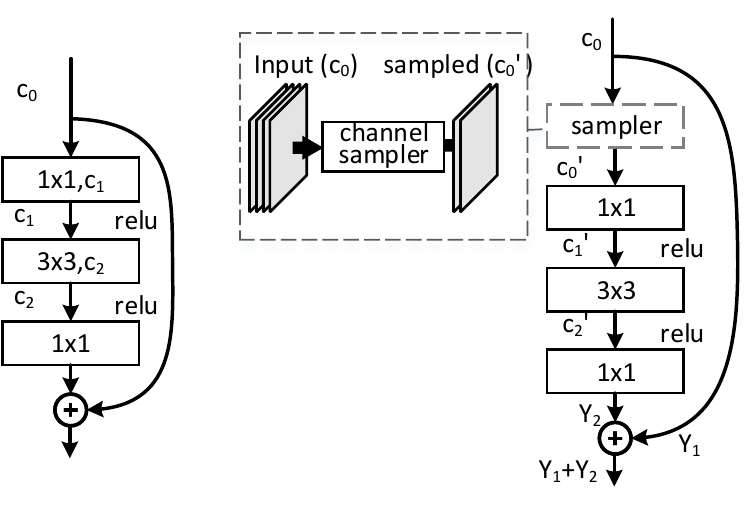}
	\caption{Illustration of multi-branch enhancement for residual block. 
		\textbf{Left}: original residual block. 
		\textbf{Right}: pruned residual block with enhancement, $\mathrm{c_x}$ denotes the \featch. Input channels of the first \conv\ are sampled, so that the large input \featch\ could be reduced. As for the last layer, rather than approximate $\reY_2$, we try to approximate $\reY_1+\reY_2$ directly (Sec.~\ref{sec:multi} Last layer of residual branch).}
	\label{fig:sampler}
	\end{figure}

	The whole model pruning discussed above is enough for single-branch networks like LeNet \cite{lecun1998gradient}, AlexNet \cite{krizhevsky2012imagenet} and VGG Nets \cite{Simonyan2014}.
	However, it is insufficient for multi-branch networks like GoogLeNet \cite{szegedy2015going} and  ResNet \cite{He2015}.
	We mainly focus on pruning the widely used residual structure (\eg, ResNet \cite{He2015}, Xception \cite{chollet2016xception}).
	Given a residual block shown in Fig.~\ref{fig:sampler}~(left), the input bifurcates into shortcut and residual branch.
	On the residual branch, there are several \conv s (\eg, 3 \conv s which have spatial size of $1\times1,3\times3,1\times1$, Fig.~\ref{fig:sampler}, left). 
	Other layers except the first and last layer can be pruned as is described previously.
	For the first layer, the challenge is that the large input \featch\ (for ResNet, 4 times of its output) can't be easily pruned, since it's shared with shortcut.
	For the last layer, accumulated error from the shortcut is hard to be recovered, since there's no parameter on the shortcut.
	To address these challenges, we propose several variants of our approach as follows.

	\label{sec:multi}
	\textbf{Last layer of residual branch}: 
	Shown in Fig.~\ref{fig:sampler}, the output layer of a residual block consists of two inputs: 
	feature map $\reY_1$ and $\reY_2$ from the shortcut and residual branch.
	We aim to recover $\reY_1 + \reY_2$ for this block.
	Here, $\reY_1, \reY_2$ are the original feature maps before pruning.
	$\reY_2$ could be approximated as in Eqn.~\ref{eq:l0}.
	However, shortcut branch is parameter-free, then $\reY_1$ could not be recovered directly.
	To compensate this error, 
	the optimization goal of the last layer is changed from $\reY_2$ to $\reY_1 - \reY^\prime_1 + \reY_2$,
	which does not change our optimization.
	Here, $\reY^\prime_1$ is the current feature map after previous layers pruned.
	When pruning, volumes should be sampled correspondingly from these two branches.

	\textbf{First layer of residual branch}: Illustrated in Fig.~\ref{fig:sampler}(left),
	the input feature map of the residual block could not be pruned,
	since it is also shared with the shortcut branch.
	In this condition, we could perform \sampling\ before the first convolution to save computation.
	We still apply our algorithm as Eqn.~\ref{eq:l0}.
	Differently, we sample the selected channels on the shared feature maps to construct a new input for the later convolution,
	shown in Fig.~\ref{fig:sampler}(right).
	Computational cost for this operation could be ignored.	
	More importantly, after introducing \textit{feature map sampling},
	the convolution is still "regular".
	
	\Filterwise\ is another option for the first convolution on the residual branch.
	Since the input channels of parameter-free shortcut branch could not be pruned,
	we apply our Eqn.~\ref{eq:l0} to each filter independently (each filter chooses its own representative input channels).
	Under single layer acceleration, \filterwise\ is more accurate than our original one.
	From our experiments, 
	it improves 0.5\% top-5 accuracy for \x{2} ResNet-50 (applied on the first layer of each residual branch) without fine-tuning.
	However, after fine-tuning, there's no noticeable improvement.
	In addition, it outputs "irregular" \conv s, which need special library implementation support.
	We do not adopt it in the following experiments.


\section{Experiment}

	\newcommand{\tworow}[1]{\begin{tabular}[c]{@{}c@{}}#1\end{tabular}}

	\newcommand{\jadercite}{Jaderberg \etal \cite{jaderberg2014speeding}}
	\newcommand{\jader}{Jaderberg \etal \cite{jaderberg2014speeding} (\cite{Zhang2015}'s impl.)}
	\newcommand{\asym}{Asym. \cite{Zhang2015}}
	\newcommand{\asymd}{Asym. 3D \cite{Zhang2015}}
	\newcommand{\asymdft}{Asym. 3D (fine-tuned) \cite{Zhang2015}}
	\newcommand{\prli}{\filterpruning\ \cite{Li2016}}
	\newcommand{\pr}{\filterpruning\ \cite{Li2016} (our impl.)}
	\newcommand{\prft}{\filterpruning\ \cite{Li2016} (fine-tuned, our impl.)}
	\newcommand{\prfttwo}{\tworow{\filterpruning\ \cite{Li2016}\\ (fine-tuned, our impl.)}}

	We evaluation our approach for the popular VGG Nets \cite{Simonyan2014}, ResNet \cite{He2015}, Xception \cite{chollet2016xception} on ImageNet \cite{JiaDeng2009}, CIFAR-10 \cite{Krizhevsky2009} and PASCAL VOC 2007 \cite{pascal-voc-2007}.
	
	For Batch Normalization \cite{ioffe2015batch}, 
	we first merge it into convolutional weights, 
	which do not affect the outputs of the networks. So that each \conv\ is followed by ReLU \cite{nair2010rectified}.
	We use Caffe \cite{jia2014caffe} for deep network evaluation,
	and scikit-learn \cite{scikit-learn} for solvers implementation.
	For \channelpruning, we found that it is enough to extract 5000 images, and 10 samples per image, which is also efficient (i.e. several minutes for VGG-16 \footnote{On Intel Xeon E5-2670 CPU}).
	On ImageNet, we evaluate the top-5 accuracy with single view. 
	Images are resized such that the shorter side is 256.
	The testing is on center crop of $224\times224$ pixels.
	We could gain more performance with fine-tuning.
	We use a batch size of 128 and learning rate $1e^{-5}$.
	We fine-tune our pruned models for 10 epoches (less than 1/10 iterations of training from scratch).
	The augmentation for fine-tuning is random crop of $224\times224$ and mirror.
	
	
	\subsection{Experiments with VGG-16}
	VGG-16 \cite{Simonyan2014} is a 16 layers single-branch convolutional neural network, with 13 \conv s. 
	It is widely used in recognition, detection and segmentation, \etc.
	Single view top-5 accuracy for VGG-16 is \origvgg\%\footnote{http://www.vlfeat.org/matconvnet/pretrained/}. 
	
	\subsubsection{Single Layer Pruning}\label{sec:ablation}\label{sec:filtersingle}

	\begin{figure*}
	\centering
	\includegraphics[width=0.8\linewidth]{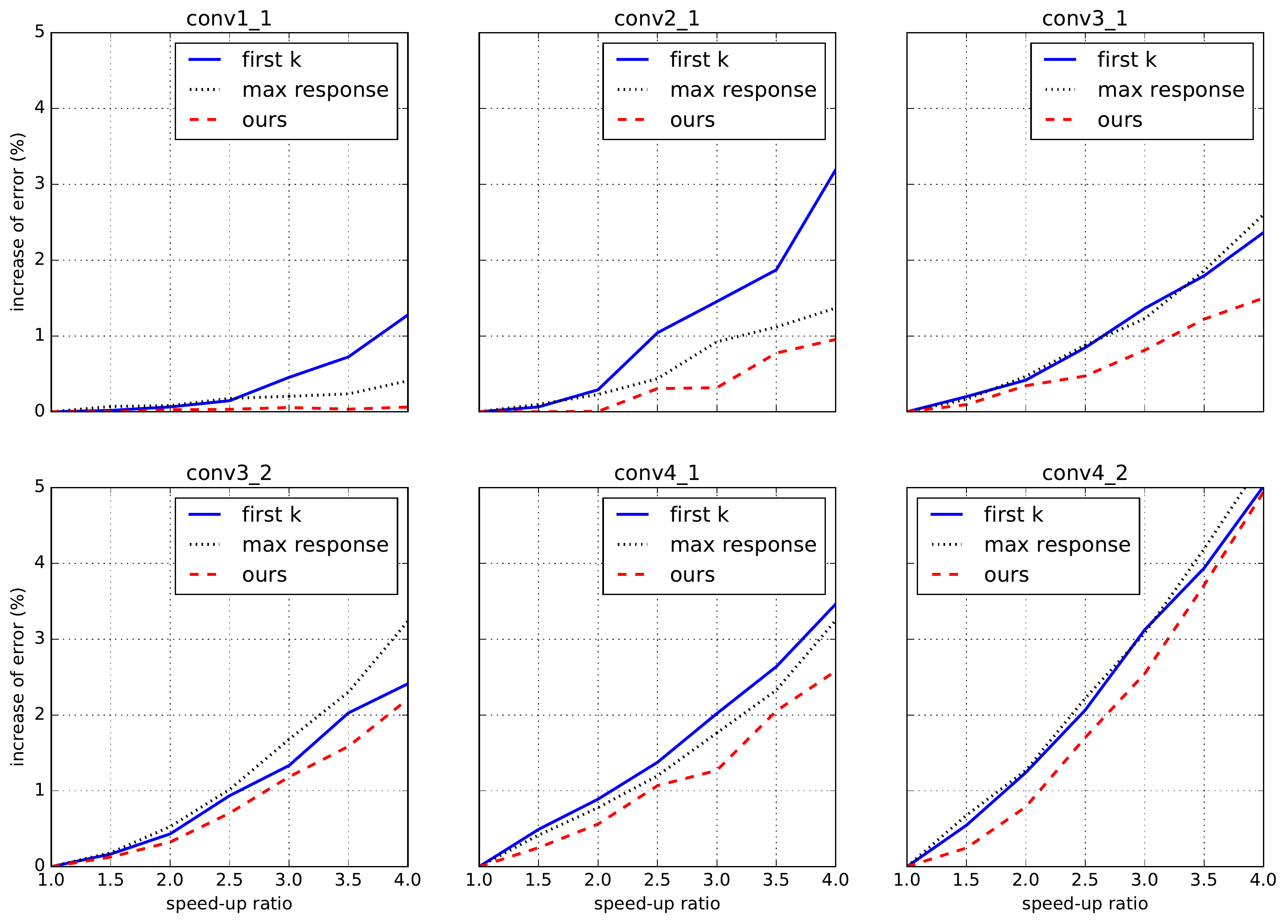}
	\caption{Single layer performance analysis under different speed-up ratios (without fine-tuning), measured by increase of error.
		To verify the importance of channel selection refered in Sec.~\ref{sec:linear}, we considered two naive baselines.
		 \firstk\ selects the first $k$ feature maps. \prune\ selects channels based on absolute sum of corresponding weights filter \cite{Li2016}. Our approach is consistently better (\textit{smaller is better}).}
	\label{fig:ablation}
	\end{figure*}
	
	In this subsection,  we evaluate single layer acceleration performance using our algorithm in Sec.~\ref{sec:linear}.
	For better understanding, we compare our algorithm with two naive channel selection strategies.
	\firstk\ selects the first \textit{k} channels. 
	\prune\ selects channels based on corresponding filters that have high absolute weights sum \cite{Li2016}.
	For fair comparison, 
	we obtain the feature map indexes selected by each of them, 
	then perform reconstruction (Sec. \ref{sec:linear} (ii)).
	We hope that this could demonstrate the importance of channel selection.
	Performance is measured by increase of error after a certain layer is pruned without fine-tuning, shown in Fig.~\ref{fig:ablation}.
	 
	As expected, error increases as \ratio\ increases.
	Our approach is consistently better than other approaches in different \conv s under different \ratio.
	Unexpectedly, sometimes \prune\ is even worse than \firstk. 
	We argue that \prune\ ignores correlations between different filters. 
	Filters with large absolute weight may have strong correlation. 
	Thus selection based on filter weights is less meaningful.
	Correlation on feature maps is worth exploiting.
	We can find that channel selection affects reconstruction error a lot. Therefore, it is important for \channelpruning.
	
	Also notice that \channelpruning\ gradually becomes hard, from shallower to deeper layers. 
	It indicates that shallower layers have much more redundancy,
	which is consistent with \cite{Zhang2015}.
	We could prune more aggressively on shallower layers in whole model acceleration.
	
	\subsubsection{Whole Model Pruning} \label{sec:ratio}
		\begin{table}
		\centering
		\begin{tabular}{|c|c|c|c|}
			\hline 
			\multicolumn{4}{|c|}{Increase of top-5 error (1-view, baseline \origvgg\%)}                              \\ \hline
			Solution & \x{2} & \x{4} & \x{5}  \\ 
			\hline 
			\jader   & -  & 9.7 & 29.7  \\ 		
			\hline 
			\asym   & 0.28  &  3.84 & -  \\ 
			\hline 
			\prfttwo & 0.8 & 8.6  & 14.6  \\ 
			\hline 
			Ours (without fine-tune) & \vggtworaw & \vggfourraw  & \vggfiveraw  \\ 
			\hline 
			Ours (fine-tuned) & 0 & \vggfour & \vggfive \\ 
			\hline	
		\end{tabular} 
		\caption{Accelerating the VGG-16 model \cite{Simonyan2014} using a speedup ratio of \x{2}, \x{4}, or \x{5} (\textit{smaller is better}).}
		\label{tab:theo}
	\end{table}

	Shown in Table~\ref{tab:theo}, whole model acceleration results under \x{2}, \x{4}, \x{5} are demonstrated.
	We adopt whole model pruning proposed in Sec.~\ref{sec:whole}.
	Guided by single layer experiments above, we prune more aggressive for shallower layers.
	Remaining channels ratios for shallow layers (\verb|conv1_x| to \verb|conv3_x|) and deep layers (\verb|conv4_x|) is $1:1.5$.
	\verb|conv5_x| are not pruned, since they only contribute 9\% computation in total and are not redundant.
	
	After fine-tuning, we could reach \x{2} speed-up without losing accuracy.
	Under \x{4}, we only suffers \vggfour\% drops.
	Consistent with single layer analysis, our approach outperforms previous \channelpruning\ approach (Li \etal \cite{Li2016}) by large margin.
	This is because we fully exploits channel redundancy within feature maps.
	Compared with \tensordecom\ algorithms, our approach is better than \jadercite,
	without fine-tuning. 
	Though worse than \asym,
	our combined model outperforms its combined Asym. 3D (Table~\ref{tab:combine}).
	This may indicate that \channelpruning\ is more challenging than \tensordecom, 
	since removing channels in one layer might dramatically change the input of the following layer.
	However, \channelpruning\ keeps the original model architecture, do not introduce additional layers, 
	and the absolute \ratio\ on GPU is much higher (Table~\ref{fig:real}).
	

	\begin{table}
	\centering
	\begin{tabular}{|c|c|c|}
		\hline 
		\multicolumn{3}{|c|}{Increase of top-5 error (1-view, \origvgg\%)}                              \\ \hline
		Solution & \x{4} & \x{5}  \\ 
		\hline 
		\asymd & 0.9 & 2.0  \\ 
		\hline
		\asymdft & 0.3 & 1.0  \\ 
		\hline
		Our 3C & \vggcfour  & \vggc  \\ 
		\hline 
		Our 3C (fine-tuned) & \textbf{\vggcftfour} & \textbf{\vggcft}  \\ 
		\hline			
	\end{tabular} 
	\caption{Performance of combined methods on the VGG-16 model \cite{Simonyan2014} using a \ratio\ of \x{4} or \x{5}. 
		Our 3C solution outperforms previous approaches (\textit{smaller is better}).}
	\label{tab:combine}
	\end{table}

	Since our approach exploits a new cardinality,
	we further combine our \channelpruning\ with spatial factorization \cite{jaderberg2014speeding} and channel factorization \cite{Zhang2015}.
	Demonstrated in Table~\ref{tab:combine}, our 3 cardinalities acceleration (spatial, channel factorization, and \channelpruning, denoted by 3C) outperforms previous state-of-the-arts.
	\asymd\ (spatial and channel factorization), factorizes a \conv\ to three parts: $1\times3,\ 3\times1,\ 1\times1$.
	
	We apply spatial factorization, channel factorization, and our \channelpruning\ together sequentially layer-by-layer. 
	We fine-tune the accelerated models for 20 epoches, 
	since they are 3 times deeper than the original ones.
	After fine-tuning, our \x{4} model suffers no degradation.
	Clearly, a combination of different acceleration techniques is better than any single one.
	This indicates that a model is redundant in each cardinality.

\subsubsection{Comparisons of Absolute Performance}\label{sec:gpu}
\begin{table*}[]
	\centering
	\begin{tabular}{|c|c|c|c|}
		\hline
		Model                      & Solution        & Increased err.               & GPU time/ms       \\ \hline
		VGG-16                      & -               & 0                     &  \round{3}{8.144125}          \\ \hline
		\multirow{5}{*}{VGG-16 (\x{4})} & \jader       & 9.7                          & \round{3}{8.05059375} (\x{1.01}) \\ \cline{2-4} 
		& \asym       & 3.8                          & \round{3}{5.243625} (\x{1.55})  \\ \cline{2-4} 
		& \asymd    & 0.9                          &  \round{3}{8.5031875} (\x{0.96})  \\ \cline{2-4} 
		& \asymdft    & \textbf{0.3}                         &  \round{3}{8.5031875} (\x{0.96})  \\ \cline{2-4} 
		& Ours (fine-tuned) & \vggfour & \textbf{3.264 (\x{2.50})}     \\ \hline 
	\end{tabular}
	\caption{GPU acceleration comparison. We measure forward-pass time per image. Our approach generalizes well on GPU (\textit{smaller is better}).}
	\label{fig:real}
\end{table*}

	We further evaluate absolute performance of acceleration on GPU.
	Results in Table~\ref{fig:real} are obtained under Caffe \cite{jia2014caffe}, 
	CUDA8 \cite{cuda} and cuDNN5 \cite{cudnn}, with a mini-batch of 32 on a GPU (GeForce GTX TITAN X).
	Results are averaged from 50 runs.
	\Tensordecom\ approaches decompose weights into too many pieces, which heavily increase overhead.
	They could not gain much absolute speed-up.
	Though our approach also encountered performance decadence, 
	it generalizes better on GPU than other approaches.
	Our results for \tensordecom\ differ from previous research \cite{Zhang2015,jaderberg2014speeding}, 
	maybe because current library and hardware prefer single large convolution instead of several small ones.
	
	\subsubsection{Comparisons with Training from Scratch}\label{sec:vggscratch}
		\begin{table}
	\centering
	\begin{tabular}{|c|c|c|}
		\hline 
		Original (acc. \origvgg\%) & Top-5 err. & Increased err.  \\ 
		\hline 
		From scratch  & \vggscratchacc  &  \vggscratcherr    \\ 
		\hline 
		From scratch (uniformed) & \vggscratchuniacc  & \vggscratchunierr     \\ 
		\hline 
		Ours & \vggfourrawacc & \vggfourraw  \\ 
		\hline
		Ours (fine-tuned) & \vggfouracc & \textbf{\vggfour}  \\ 
		\hline
	\end{tabular} 
	
	\caption{Comparisons with training from scratch, under \x{4} acceleration. Our fine-tuned model outperforms scratch trained counterparts (\textit{smaller is better}).}
	\label{tab:orig}
	\end{table}

	Though training a compact model from scratch is time-consuming (usually 120 epoches),
	it worths comparing our approach and from scratch counterparts.
	To be fair, we evaluated both from scratch counterpart, 
	and normal setting network that has the same computational complexity and same architecture.

	Shown in Table~\ref{tab:orig}, 
	we observed that it's difficult for from scratch counterparts to reach competitive accuracy.
	our model outperforms from scratch one.
	Our approach successfully picks out informative channels and constructs highly compact models.  
	We can safely draw the conclusion that the same model is difficult to be obtained from scratch.
	This coincides with architecture design researches \cite{huang2016speed,Alvarez2016} 
	that the model could be easier to train if there are more channels in shallower layers.
	However, \channelpruning\ favors shallower layers.
	
	For from scratch (uniformed),
	the filters in each layers is reduced by half (eg. reduce \verb|conv1_1| from 64 to 32).
	We can observe that normal setting networks of the same complexity couldn't reach same accuracy either.
	This consolidates our idea that there's much redundancy in networks while training.
	However, redundancy can be opt out at inference-time.
	This maybe an advantage of inference-time acceleration approaches over training-based approaches.
	
	Notice that there's a 0.6\% 
	gap between the from scratch model and uniformed one,
	which indicates that there's room for model exploration.
	Adopting our approach is much faster than training a model from scratch, even for a thinner one.
	Further researches could alleviate our approach to do thin model exploring.

	\subsubsection{Acceleration for Detection}    
	VGG-16 is popular among object detection tasks \cite{fasterrcnn,yolo,ssd}.
    We evaluate transfer learning ability of our \x{2}/\x{4} pruned VGG-16,
	for Faster R-CNN \cite{fasterrcnn} object detections.
	PASCAL VOC 2007 object detection benchmark \cite{pascal-voc-2007} contains 5k trainval images and 5k test images.
	The performance is evaluated by mean Average Precision (mAP) and mmAP (primary challenge metric of COCO \cite{lin2014microsoft}).
	In our experiments, we first perform \channelpruning\ for VGG-16 on the ImageNet. 
	Then we use the pruned model as the pre-trained model for Faster R-CNN.

		\begin{table}
	\centering
	\begin{tabular}{|c|c|c|c|c|}
		\hline 
		Speedup & mAP & $\Delta$ mAP & mmAP & $\Delta$ mmAP \\ 
		\hline 
		Baseline & 68.7  &  - & 36.7 & - \\ 
		\hline 
		\x{2} &  68.3 & 0.4 & 36.7 & 0.0 \\ 
		\hline 
		\x{4} & 66.9  & 1.8 & 35.1 & 1.6 \\ 
		\hline 
	\end{tabular} 
	
	\caption{\x{2}, \x{4} acceleration for Faster R-CNN detection. mmAP is AP at IoU=.50:.05:.95 (primary challenge metric of COCO \cite{lin2014microsoft}). }
	\label{tab:det}
\end{table}

	The actual running time of Faster R-CNN is 220ms / image. 
	The \conv s contributes about 64\%.
    We got actual time of 94ms for \x{4} acceleration.
   From Table~\ref{tab:det}, we observe 0.4\% mAP drops of our \x{2} model, which is not harmful for practice consideration. Observed from mmAP, For higher localization requirements our speedup model does not suffer from large degradation.
	
	\subsection{Experiments with Residual Architecture Nets}
	For Multi-path networks \cite{szegedy2015going,He2015,chollet2016xception}, 
	we further explore the popular ResNet \cite{He2015} and latest Xception \cite{chollet2016xception}, 
	on ImageNet and CIFAR-10.
	Pruning residual architecture nets is more challenging. 
	These networks are designed for both efficiency and high accuracy.
	\Tensordecom\ algorithms \cite{Zhang2015,jaderberg2014speeding} have difficult to accelerate these model.
	Spatially, $1\times1$ convolution is favored, which could hardly be factorized.
	
	\subsubsection{ResNet Pruning} \label{sec:resnetimagenet}

	\begin{table}[]
		\centering
		\begin{tabular}{|c|c|}
			\hline
			Solution                                                                                & Increased err. \\ \hline
			Ours                                                                                    & 8.0           \\  \hline
			\begin{tabular}[c]{@{}c@{}}Ours\\ (enhanced)\end{tabular}              & \resmb           \\ \hline
			\begin{tabular}[c]{@{}c@{}}Ours \\ (enhanced, fine-tuned)\end{tabular} & \resft           \\ \hline
		\end{tabular}
		\caption{\x{2} acceleration for ResNet-50 on ImageNet, the baseline network's top-5 accuracy is \resorig\% (one view). 
		We improve performance with multi-branch enhancement (Sec. \ref{sec:multi}, \textit{smaller is better}).}
		\label{tab:resnet}
	\end{table}

	ResNet complexity uniformly drops on each residual block.
	Guided by single layer experiments (Sec. \ref{sec:ablation}),
	we still prefer reducing shallower layers heavier than deeper ones. 
	
	Following similar setting as \filterpruning\ \cite{Li2016},
	we keep 70\% channels for sensitive residual blocks (\verb|res5| and blocks close to the position where spatial size change, \eg \verb|res3a,res3d|). 
	As for other blocks, we keep 30\% channels.
	With multi-branch enhancement, we prune \verb|branch2a| more aggressively within each residual block.
	The preserving channels ratios for \verb|branch2a,branch2b,branch2c| is $2:4:3$ (\eg, Given 30\%, we keep 40\%, 80\%, 60\% respectively).
	
	We evaluate performance of multi-branch variants of our approach (Sec. \ref{sec:multi}).
	From Table~\ref{tab:resnet}, we improve 4.0\% 
	with our multi-branch enhancement.
	This is because we accounted the accumulated error from shortcut connection which could broadcast to every layer after it.
	And the large input \featch\ at the entry of each residual block is well reduced by our \sampling.

   \subsubsection{Xception Pruning}

   	   \begin{table}
   		\centering
   		\begin{tabular}{|c|c|}
   			\hline
   			Solution                      & Increased err.        \\ \hline
   			\pr                                                                     & \xceptionorig        \\ \hline
   			\prfttwo &  \xceptionpr          \\ \hline
   			Ours              & \xceptioncr          \\ \hline
   			Ours (fine-tuned)       &\textbf{\xceptionft} \\ \hline
   		\end{tabular}
   	\caption{Comparisons for \xceptionfifty, under \x{2} acceleration ratio. The baseline network's top-5 accuracy is \xceptionorig\%. Our approach outperforms previous approaches. Most \structured\ methods are not effective on Xception architecture (\textit{smaller is better}).}
   	\label{tab:xception}
   \end{table}    
   Since computational complexity becomes important in model design, 
   separable convolution has been payed much attention \cite{xie2016aggregated,chollet2016xception}.
   Xception \cite{chollet2016xception} is already spatially optimized and \tensordecom\ on $1\times1$ \conv\ is destructive.
   Thanks to our approach, it could still be accelerated with graceful degradation.
   For the ease of comparison, we adopt Xception convolution on ResNet-50, denoted by Xception-50.
   Based on ResNet-50, we swap all \conv s with spatial conv blocks.
   To keep the same computational complexity, 
   we increase the input channels of all \verb|branch2b| layers by \x{2}.
   The baseline Xception-50 has a top-5 accuracy of \xceptionorig\% and complexity of 4450 MFLOPs.
   
   We apply multi-branch variants of our approach as described in Sec. \ref{sec:multi}, 
   and adopt the same pruning ratio setting as ResNet in previous section.
   Maybe because of Xception block is unstable,
   Batch Normalization layers must be maintained during pruning.
   Otherwise it becomes nontrivial to fine-tune the pruned model.

   Shown in Table~\ref{tab:xception},
   after fine-tuning, we only suffer \textbf{\xceptionft\%} increase of error under \x{2}.
   \prli\ could also apply on Xception, though it is designed for small \ratio.
   Without fine-tuning, top-5 error is 100\%. 
   After training 20 epochs which is like training from scratch, increased error reach \xceptionpr\%.
   Our results for Xception-50 are not as graceful as results for VGG-16, 
   since modern networks tend to have less redundancy by design.
   
   	\subsubsection{Experiments on CIFAR-10}
   	Even though our approach is designed for large datasets,
   	it could generalize well on small datasets.
   	We perform experiments on CIFAR-10 dataset \cite{Krizhevsky2009}, which is favored by many acceleration researches.
   	It consists of 50k images for training and 10k for testing in 10 classes. 

%

   \begin{table}[]
	\centering
	\begin{tabular}{|c|c|}
		\hline
		Solution & Increased err. \\ \hline
		\prfttwo & 1.3 \\ \hline
		From scratch  & 1.9 \\ \hline
		Ours  & 2.0 \\ \hline
		Ours (fine-tuned) & \textbf{1.0} \\ \hline
	\end{tabular}
	\caption{\x{2} speed-up comparisons for ResNet-56 on CIFAR-10, the baseline accuracy is 92.8\% (one view). We outperforms previous approaches and scratch trained counterpart (\textit{smaller is better}). 
	}
	\label{tab:cifar}
	\end{table}

   	We reproduce ResNet-56, which has accuracy of 92.8\% (Serve as a reference, the official ResNet-56 \cite{He2015} has accuracy of 93.0\%).
   	For \x{2} acceleration, we follow similar setting as Sec.~\ref{sec:resnetimagenet} (keep the final stage unchanged, where the spatial size is $8\times8$).
   	Shown in Table~\ref{tab:cifar}, our approach is competitive with scratch trained one, without fine-tuning,
   	under 
   	\x{2} speed-up.
   	After fine-tuning, our result is significantly better than \filterpruning\ \cite{Li2016} and scratch trained one.

\section{Conclusion}
To conclude, current deep CNNs are accurate with high inference costs.
In this paper, we have presented an inference-time \channelpruning\ method for very deep networks.
The reduced CNNs are inference efficient networks while maintaining accuracy,
and only require off-the-shelf libraries.
Compelling speed-ups and accuracy are demonstrated for both VGG Net and ResNet-like networks
on ImageNet, CIFAR-10 and PASCAL VOC.

In the future, we plan to involve our approaches into training time, 
instead of inference time only, 
which may also accelerate training procedure.

{\small
\bibliographystyle{ieee}
\bibliography{egbib}
}

\clearpage

\end{document}